\documentclass{article}
\usepackage{ijcai25}

\usepackage{times}
\usepackage{soul}
\usepackage{url}
\usepackage[hidelinks]{hyperref}
\usepackage[utf8]{inputenc}
\usepackage[small]{caption}
\usepackage{graphicx}
\usepackage{amsmath}
\usepackage{amsthm}
\usepackage{booktabs}
\usepackage{algorithm}
\usepackage{algorithmic}
\usepackage{multirow}
\usepackage{array}
\usepackage{amsmath}
\usepackage{amssymb}
\usepackage[switch]{lineno}

\title{Semantic-Space-Intervened Diffusive Alignment for Visual Classification}

\author{
Zixuan Li$^1$
\and
Lei Meng$^1$\thanks{Corresponding author}
\and
Guoqing Chao$^2$
\and
Wei Wu$^1$
\and
Yimeng Yang$^1$
\and
\\
Xiaoshuo Yan$^1$
\and
Zhuang Qi$^1$
\and
Xiangxu Meng$^1$
\\
\affiliations
$^{1}$School of Software, Shandong University, Jinan, China\\
$^2$School of Computer Science and Technology, Harbin Institute of
Technology, Weihai, China\\
\emails
\{lizixuan0707, wu\_wei, yanxiaoshuo, y\_yimeng, z\_qi \}@mail.sdu.edu.cn,
\\
\{lmeng, mxx\}@sdu.edu.cn,
\ guoqingchao@hit.edu.cn
}

\begin{document}

\maketitle

\begin{abstract}

Cross-modal alignment is an effective approach to improving visual classification. Existing studies typically enforce a one-step mapping that uses deep neural networks to project the visual features to mimic the distribution of textual features. However, they typically face difficulties in finding such a projection due to the two modalities in both the distribution of class-wise samples and the range of their feature values. To address this issue, this paper proposes a novel \textbf{Se}mantic-Space-Intervened \textbf{D}iffusive \textbf{A}lignment method, termed {\it SeDA}, models a semantic space as a bridge in the visual-to-textual projection, considering both types of features share the same class-level information in classification. More importantly, a bi-stage diffusion framework is developed to enable the progressive alignment between the two modalities.
Specifically, SeDA first employs a Diffusion-Controlled Semantic Learner to model the semantic features space of visual features by constraining the interactive features of the diffusion model and the category centers of visual features. In the later stage of SeDA, the Diffusion-Controlled Semantic Translator focuses on learning the distribution of textual features from the semantic space. Meanwhile, the Progressive Feature Interaction Network introduces stepwise feature interactions at each alignment step, progressively integrating textual information into mapped features. 
Experimental results show that SeDA achieves stronger cross-modal feature alignment, leading to superior performance over existing methods across multiple scenarios.

% The Diffusion-controlled Structural Consistency Constraint module 通过对扩散模型预测特征和视觉特征类别中心的约束学习视觉特征的结构信息，然后在扩散模型后期通过Diffusion-controlled Cross-modal Transformation模块学习文本特征的分布。Feature Interaction and Reconstruction Network introduces stepwise feature interactions at each alignment step, progressively injecting textual information into the visual features.

% Specifically, SeDA consists of three key modules, where the Diffusion-controlled Structural Consistency Constraint(这里先给全称) module maintains visual structural consistency by (做了什么要说出来，不能只说有什么作用，不然太虚了). The DCT(这里先给全称) module xxx (做了什么) to facilitate the smooth transfer of cross-modal features. The FIRN(这里先给全称) module introduces stepwise feature interactions at each alignment step, progressively injecting textual information into the visual features, thereby enhancing the cross-modal alignment process. Experimental results show that SeDA achieves stronger cross-modal feature alignment, leading to superior performance over existing methods across multiple scenarios.

\end{abstract}

\section{Introduction}

Cross-modal alignment aims to integrate information from different modalities to capture semantic relationships within complex data \cite{cross-modal1,cross}. It utilizes more discriminative textual representations to enhance visual classification, effectively mitigating biases caused by the diversity of visual data, lighting conditions, and background noise \cite{noise,noise2,noise3}. Although cross-modal alignment generally outperforms single-modal learning, its effectiveness can decline due to semantic ambiguities between modalities \cite{ATNet,align}. This is primarily due to substantial differences in semantics, structures, or representational forms across modalities, which pose significant challenges for cross-modal alignment in handling high heterogeneity.

% --------------motivation---------------

% 常见的特征对齐方法以及所提出的 DMANet 对齐框架。在（a）中，传统的特征对齐过程导致视觉特征分布变化很小，并且对齐的视觉特征仍然没有完全学习到视觉特征和文本特征的语义信息。在（b）中，DMANet 通过两阶段的对齐方法，逐步将视觉特征转移到文本特征、揭示视觉模态中隐藏的文本信息并集成这些信息来增强图像分类学习，从而减轻异质性。

% 常见的特征对齐方法以及所提出的 SeDA 对齐框架。在（a）中，传统的特征对齐过程无法学习到文本特征的底层分布，且类间仍存在混淆。在（b）中，SeDA 通过渐进式的对齐方法，通过两阶段的学习，逐步将视觉特征转移到文本特征空间。

\begin{figure}[t]
\begin{center}
% [width=0.5\textwidth,height=0.5\textwidth]
\includegraphics[width=0.48\textwidth]{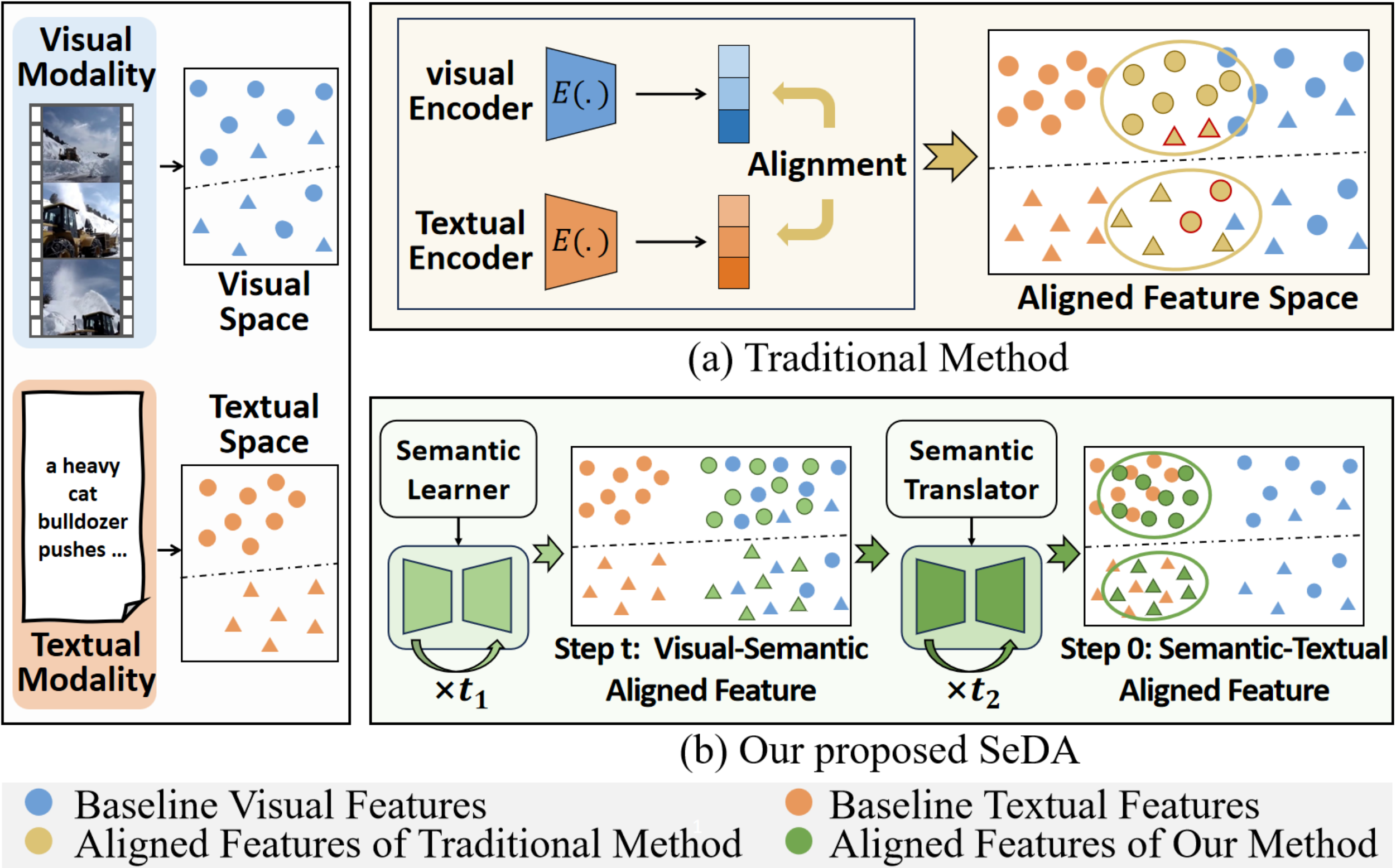}
\end{center}
\vspace{-0.3cm}
   \caption{Common feature alignment methods and the proposed SeDA alignment framework. In (a), traditional feature alignment processes fail to capture the underlying distribution of textual features, resulting in persistent inter-class confusion. In (b), SeDA employs a semantic-space-intervened diffusive alignment method, transferring visual features to the textual features space step by step through a bi-stage learning process.}
\label{fig:motivation}
\vspace{-0.3cm}

\end{figure}

To alleviate modality heterogeneity, existing methods can be broadly divided into two groups: distance metric-based alignment methods and knowledge distillation-based alignment methods. The former approach uses techniques like Maximum Mean Discrepancy (MMD) \cite{MMD}, Correlation Alignment (CORAL) \cite{coral}, and the Wasserstein Distance \cite{SWD,WD2} to explicitly minimize the differences in feature distributions or representations between modalities \cite{qizhuang}. The latter approach leverages  \cite{C2KD,MM-PKD,MoMKE} to transfer knowledge from the teacher modality to the student modality, enabling different modalities to generate similar feature representations.
% However, as shown in Figure \ref{fig:motivation}(a), due to significant distribution differences between modalities, directly bridging the gap between visual and textual features via a direct one-step mapping often fails to capture the underlying distribution of text features effectively. 
However, as shown in Figure \ref{fig:motivation}(a), due to the significant differences in class-wise sample distributions and feature value ranges between modalities, directly applying a one-step mapping often fails to find an effective projection, failing to capture the underlying distribution of textual features fully.
Furthermore, the complex relationships between modalities make it challenging for one-step alignment methods to adequately address modality-specific distributions that are highly correlated with each modality’s intrinsic discriminability.
Consequently, semantic ambiguity remains, leading to significant inter-class confusion in the aligned features.

To address the above challenges, we propose a \textbf{Se}mantic-Space-Intervened \textbf{D}iffusive \textbf{A}lignment (SeDA) method, as illustrated in Figure \ref{fig:motivation}(b). SeDA leverages the Markov reverse process of diffusion models to smoothly learn distributions over multiple steps, focusing on the semantic consistency of multimodal features. This approach alleviates the semantic ambiguity caused by the heterogeneity of textual and visual features, which arises from direct one-step mapping.
SeDA consists of three key modules: the Diffusion-controlled Semantic Learner (DSL), the Diffusion-controlled Semantic Translator (DST), and the Progressive Feature Interaction Network (PFIN). 
Specifically, we propose a Bi-stage optimization framework that models a modality-shared semantic space as an intermediary to enable a three-stage projection: from visual space to semantic space and then to textual space.
In the early stage of the diffusion process, the DSL module progressively removes irrelevant low-level visual information by regularizing the distance between the features learned from the PFIN module and the category center of the original visual features, thereby modeling semantic space from visual representations. In the later stage, the DST module guides the transformation of semantic space to the textual representation space by measuring the distributional differences between semantic and textual features.
Additionally, the PFIN module designs a diffusion network based on a cross-attention mechanism for multimodal feature fusion, ensuring the gradual introduction of textual information into mapped features through interaction at each diffusion step.
SeDA not only ensures similarity between the mapped feature distribution and the textual distribution but also considers modality-independent semantic information in both visual and textual representations.

% ----------------第四段------------------
% 本文在不同场景中验证了分类的有效性，包括通用数据集NUS-WIDE、领域数据集VIREO172和视频数据集MSRVTT。并进行性能比较、关键模块的消融研究、案例研究关键区域等实验。实验结果表明，DMANet利用文本描述的可判别性来改善视觉特征的分布，缓解了模态之间的异构性，实现视觉特征向文本特征的转换。In summary, the main contributions of this paper include:
% Extensive experiments were conducted on two OOD generalization datasets, including performance comparisons, ablation studies of key modules, case studies with visual attention visualizations, showcasing the association between background and labels. The results show that FedDDL effectively mitigates the interference of background and focuses on the objects of samples in unseen environments, improving the effectiveness of collaborative learning. To summarize, this paper presents three key contributions

Extensive experiments are conducted on the general dataset NUS-WIDE, the domain-specific dataset VIREO Food-172, and the video dataset MSRVTT, including performance comparisons, ablation studies, in-depth analysis, and case studies.
% This paper validates the effectiveness of the classification in different scenarios, including the general dataset NUS-WIDE, the domain-specific dataset VIREO Food-172, and the video dataset MSRVTT. Experiments such as performance comparisons, ablation studies, and case studies of critical regions are conducted. 
The experimental results show that SeDA models a semantic space as a bridge for the visual-to-textual projection, alleviating modality heterogeneity and achieving the transformation from visual features to textual features. The contributions of this paper are as follows:

% 1.本文提出了一种渐进式扩散对齐范式，称为ProDiffAlign用于实现缓解模态之间的异构性，据我们所知，DMANet是第一篇使用扩散模型进行对齐的工作。

% 2.所提出的基于扩散模型的渐进式对齐方法是一个即插即用的组件，可以轻松集成到各种方法中，它实现了在保留视觉模型结构信息的同时有效提取文本模态中的更多语义信息。
% 2.它利用多模态特征的语义一致性实现从视觉特征到文本特征的有效映射。

% 3.实验结果表明，通过扩散模型可以学习到文本特征的底层分布，实现视觉特征向文本特征的迁移，缓解异构特征之间的差异性。这为今后的研究提供了一种可行的途径。

\begin{itemize}
    \item A novel framework SeDA is proposed to enable the alignment between visual and textual features by modeling a diffusion process. To the best of our knowledge, SeDA is the first work to use diffusion models for cross-modal alignment in classification.

    \item The developed diffusion process improves the visual-to-textual feature projection by modeling a semantic space, which may capture higher-level semantic relationships between visual and textual representations, serving as an ``intermediary layer'' that effectively reduces the heterogeneity between different modalities. 
    
    % \item SeDA has been verified to be a model-agnostic framework that can integrate into various visual backbones. It leverages the semantic consistency of multimodal features to achieve an effective mapping from visual features to textual features.

    \item SeDA is a model-agnostic framework that can integrate into various visual backbones. It effectively learns the underlying distribution of textual features, providing a feasible approach for future research.

    % \item Experimental results demonstrate that the diffusion model effectively learns the underlying distribution of textual features, enabling the migration of visual features to textual features. This provides a feasible approach for future research.
\end{itemize}

\section{Related Work}

% 相关工作扩散模型写一节，跨模态对齐写一节？

%为了缓解异构模态之间的差异性，现有的研究一般可以分为两大类，包括基于特征差异度量的方法和基于知识蒸馏的方法。

\subsection{Cross-Modal Alignment}

\textbf{Distance Metrics-based Alignment Methods} focus on minimizing or maximizing metrics to bring data from different modalities into a common feature or decision space. Common metrics include Euclidean distance, cosine similarity, and covariance differences. For instance, Coral \cite{coral} aligns source and target features by minimizing the Frobenius norm of their feature differences. Deep Coral \cite{deep_coral} extends this by incorporating decision-level information and utilizing covariance matrices as a new alignment metric. Similarly, CLIP \cite{CLIP}, ECRL \cite{ECRL}, and TEAM \cite{TEAM} enhance cross-modal alignment by computing cosine similarity between visual and textual representations, effectively bridging the gap between modalities in the shared space.

% \subsubsection{Methods based on Knowledge Distillation}

\noindent \textbf{Knowledge Distillation-based Alignment Methods} transfer knowledge from one modality to another by employing a pre-trained teacher model to guide a student model. This ensures that both modalities generate similar representations in a shared alignment space. For example, C2KD \cite{C2KD} utilizes bidirectional distillation and dynamically filters samples with misaligned soft labels to improve alignment. MM-PKD \cite{MM-PKD} employs a multimodal teacher network to guide an unimodal student network through joint cross-attention fusion. Furthermore, PKDOT \cite{PKDOT} leverages entropy-regularized optimal transport to distill structural knowledge, enhancing stability and robustness in the multimodal distillation process.

\subsection{Diffusion Models}

Diffusion models are inspired by non-equilibrium thermodynamics. DDPM \cite{ddpm} gradually adds noise to the data distribution and trains a neural network to learn the reverse diffusion process, thereby denoising images corrupted by Gaussian noise. Most diffusion model studies focus on generative tasks, such as image generation \cite{wang2023anything}, 3D content generation \cite{dreamfont3d}, and video generation \cite{ddpm_video}. Recently, some work has applied diffusion models to discriminative tasks \cite{MMRec}. For example, DiffusionDet \cite{ddpm_dec} formulates object detection as a diffusion denoising process, progressively refining noisy bounding boxes into object boxes.  DiffusionRet \cite{ddpm_ret} models the correlation between text and video as their joint probability and approaches retrieval as a gradual generation of this joint distribution from noise. DiffuMask \cite{diffumask} uses text-guided cross-attention information to locate class- or word-specific regions, resulting in semantic masks for synthesized images.

\begin{figure*}[t]
\centering
\includegraphics[width=0.96\textwidth]{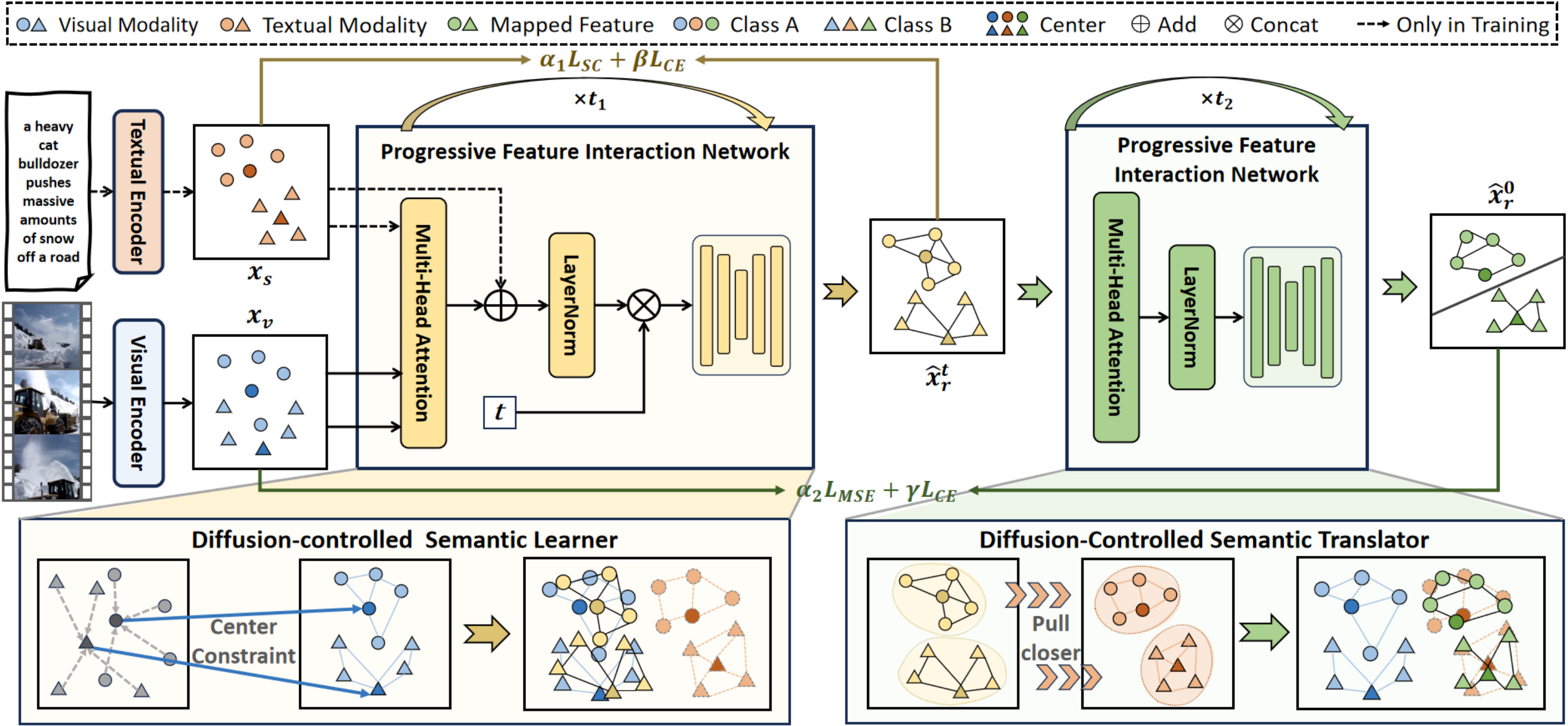}
% \vspace{-0.2cm}
\caption{Illustration of the proposed SeDA. SeDA takes visual-textual data pairs as input, which are processed by dedicated neural networks for vision and text to extract global features $x_v$ and $x_s$. The PFIN module progressively integrates textual information, while the DSL and DST modules work together to align visual and textual features effectively.} 
% \vspace{-0.4cm}
\label{fig:framework}
\end{figure*}

\section{Problem Formulation}

% 是否要加一个介绍扩散模型原理？
% \subsection{Problem Formulation}
% 我们的研究目标是通过使用特权信息学习（LUPI，Learning Using Privileged Information）来增强跨模态对齐网络的分类性能。该方法旨在从配对的图像-文本数据中提取更全面和更丰富的视觉表示。具体来说，数据表示为 $Data={P,T}$，其中稀缺但宝贵的文本数据 $T$ 作为特权信息用于指导图像数据 $P$ 的表示。训练样本由 $N$ 对图像-文本数据组成，即 $S_N={(p_1,t_1),(p_2,t_2),...,(p_n,t_n)}$，而测试样本仅包含 $M$ 张图像，即 $S_M={p_1,p_2,...,p_m}$。我们使用视觉编码器 $E_v$ 提取图像表示 $R_v={x_v^1,x_v^2,...,x_v^n}$，其中 $x_v=E_v(P)$。类似地，对于文本数据，表示为 $R_s={x_s^1,x_s^2,...,x_s^n}$，其中 $x_s=E_s(T)$。研究将图像表示 $x_v$ 和文本表示 $x_s$ 作为后续模块的输入，并最终结合它们以获得预测结果 $\hat{C}$，并根据特定指标对模型进行评估。

The goal of this research is to improve the classification performance of cross-modal alignment networks by leveraging learning using privileged information (LUPI), where textual data is available only during the training phase. The dataset is composed of paired visual data $V$ and textual data $T$, with the rare but informative textual data $T$ serving as privileged information to enhance the representation of visual data $V$. Specifically, the training set consists of $N$ visual-textual pairs $D_N={(v_1,t_1),(v_2,t_2),...,(v_n,t_n)}$, while the test set contains only $M$ visual samples $D_M={v_1,v_2,...,v_m}$.
A visual encoder $E_v$ is employed to extract visual features $x_v=E_v(V)$, while a textual encoder $E_t$ processes the textual data to extract features $x_t=E_t(T)$. The extracted visual and textual representations are aligned in subsequent modules. Finally, the aligned features are fed into a classifier to produce the prediction $\hat{C}$, with model performance evaluated using specific metrics.

% \subsection{Denoising Diffusion Probabilistic Models (DDPM)}

% The denoising diffusion probabilistic model \cite{ddpm} includes a forward process $q$ and a reverse process $p$. In the forward process, Gaussian noise is added step-by-step to the data $x_0$, following a Markov process with a predefined variance schedule $\{\beta_i\}_{i=1}^T$:
% \begin{align}
% q\left(x_i \mid x_0\right) &= \mathcal{N}\left(x_i ; \sqrt{\bar{\alpha}_i} x_0,\left(1-\bar{\alpha}_i\right) \mathbf{I}\right)
% \label{eq:forward}
% \end{align}
% where $x_0 \sim q(x)$, $\mathcal{N(.)}$ means a Gaussian distribution. $\alpha_i = 1-\beta_i$ and $\bar{\alpha}_i = \prod_{s=0}^i \alpha_s$, utilize the above formula to sample the noisy sample $x_i$ at any step $i$ from $x_0$.

% In the reverse process, the goal is to iteratively reconstruct the original data by optimizing the likelihood $p_\theta(x_0)$. The reverse process is defined by:
% \begin{align}
% p_\theta\left(x_{i-1} \mid x_i\right) &= \mathcal{N}\left(x_{i-1} ; \mu_\theta\left(x_i, i\right), \tilde{\beta}_i \mathbf{I} \right)
% \end{align}
% where $\mu_{\theta}$ is the predicted mean, and $\tilde{\beta}_i = \frac{1 - \bar{\alpha}_{i-1}}{1 - \bar{\alpha}_i} \beta_i$ is the variance term. The formula for $\mu_{\theta}$ is given by:

% \begin{align}
%     \mu_\theta\left(x_i, i\right) &= \frac{1}{\sqrt{\alpha_i}}\left(x_i-\frac{\beta_i}{\sqrt{1-\bar{\alpha}_i}} \epsilon_\theta\left(x_i, i\right)\right)
%     \label{eq:mu}
% \end{align}
% where $\epsilon_{\theta}(x_i,i)$ is a neural network that predicts the noise at each step.

\section{Method}

%本研究提出了一种基于扩散模型（DMANet）的跨模态对齐方法，以减轻不同模态之间的异质性，如图\ref{fig:framework}所示。 DMANet 由三个主要模块组成：扩散控制的结构一致性对齐（DSCA）模块、扩散控制的跨模态变换（DCT）模块和扩散特征交互与重建网络（FIRN）模块。 DSCA模块在扩散模型重建的早期阶段从原始视觉分布中学习结构信息，增强模型对视觉信息关键区域的感知。 DCT 模块在扩散模型重建的后期阶段从文本信息中学习语义内容。 FFRN模块通过重新设计扩散模型的网络结构，促进视觉和文本信息之间的深度交互。以下是这些模块的详细信息。

% %本研究提出了一种Semantic-space-guided diffusive alignment method (SeDA)，将模态共享的语义空间作为媒介，实现视觉表征向文本表征的映射，以减轻不同模态之间的异质性，如图\ref{fig:framework}所示。 SeDA 由三个主要模块组成：Progressive Feature Interaction Network (PFIN)模块，Diffusion-controlled Semantic  Extractor（DSE） 模块和Diffusion-controlled semantic translator（DST）。 DSE模块在扩散模型的早期阶段从原始视觉分布中提取模态无关的语义特征。 DST 模块在扩散模型的后期阶段由语义特征空间向文本特征空间映射。 PFIN模块通过重新设计扩散模型的网络结构，实现文本信息的逐步引入。以下是这些模块的详细信息。

 This study proposes a Semantic-Space-Intervened Diffusive Alignment method (SeDA), which utilizes a modality-shared semantic space as an intermediary to enable the mapping from visual representations to textual representations, thereby effectively mitigating the heterogeneity between different modalities, as shown in Figure \ref{fig:framework}. SeDA consists of three primary modules: the Progressive Feature Interaction Network (PFIN), the Diffusion-controlled Semantic Learner (DSL), and the Diffusion-controlled Semantic Translator (DST). The DSL module operates in the early stage of the diffusion model, learning modality-independent semantic space from the original visual distribution. The DST module works in the later stage, projecting the semantic space to the textual features space. The PFIN module progressively introduces textual information by redesigning the diffusion model's network structure. Details of these modules are described below.

% This study proposes a Semantic-space-guided diffusive alignment method (SeDA) to alleviate the heterogeneity between different modalities, as shown in Figure \ref{fig:framework}. SeDA consists of three main modules: the Diffusion-controlled Structural Consistency Constraint (DSCC) module, the Diffusion-controlled Cross-modal Transformation (DCT) module, and the Feature Interaction and Reconstruction Network (FIRN) module. The DSCC module learns structural information from the original visual distribution during the early stages of diffusion model, enhancing the model's perception of key areas in visual information. The DCT module learns semantic content from text information during the later stages of diffusion model. The FFRN module facilitates deep interaction between visual and textual information by redesigning the network structure of the diffusion model. The following are the details of these modules.

\subsection{Progressive Feature Interaction Network (PFIN)}

% % 在 Diffusive Feature Fusion and Reconstruction Network (FFRN) 模块中，我们利用跨模态注意力机制和多模态特征融合来实现视觉信息和文本信息之间的深度交互。该模块确保视觉信息能够有效地与文本特征融合和对齐，从而增强模型的跨模态表示能力。

% % 首先通过跨模态注意力机制，视觉和文本特征在潜在空间中被有效地结合，以捕捉视觉信息与文本语义之间的深层关系。具体来说，给定视觉特征 $x_v$ 和文本特征 $x_s$，通过跨模态注意力机制生成融合特征表示 $F_{attn}$，其过程可表示为：
% $$ \mathbf{F}_{t} = \text{softmax}\left(\frac{Q K^T}{\sqrt{d_k}}\right) V $$
% % 其中，文本特征$x_s$作为查询（Query），图像特征$x_v$作为键（Key）和值（Value）

% % 使用正弦位置编码函数 $\text{SinusoidalPosEmb}(t)$ 对时间步 $t$ 进行嵌入，得到时间特征向量 $\mathbf{e}_{t}$：
% \begin{equation}    
% \mathbf{e}_{t} = \phi(\mathbf{W}_{2} \cdot \phi(\mathbf{W}_{1} \cdot \text{SinusoidalPosEmb}(t))),
% \end{equation}
% % 其中，$\mathbf{W}{1}$ 和 $\mathbf{W}{2}$ 为权重矩阵，$\phi$ 为激活函数。

% % 随后，我们将注意力模块的输出F_{t}与编码后的扩散时间特征$\mathbf{e}_{t}$连接起来作为去噪解码器的输入数据:
% \begin{equation}
%     \mathbf{h}_{0} = [\mathbf{F}_{t}; \mathbf{e}_{t}]
% \end{equation}

% % 去噪解码器是一个多层感知器（MLP），包含一个具有多层线性变换和激活函数的中间层用于编码特征，以及一个用于计算输出分布的线性层。

% In the Diffusive Feature Interaction and Reconstruction Network module, we leverage cross attention mechanisms and multimodal feature fusion to achieve deep interaction between visual information and textual information. This module ensures that visual information can be effectively integrated and aligned with textual features, thereby enhancing the model's cross-modal representation capability.

In the Progressive Feature Interaction Network module, we use cross-attention mechanisms and multimodal feature fusion to enable deep interaction between visual and textual information. This ensures effective integration and alignment of visual and textual features, enhancing cross-modal representation.

% Using the cross attention mechanism, visual and textual features are fused in the latent space to capture their deep relationship. 

Unlike vanilla diffusion models \cite{ddpm}, which predict $\epsilon=\epsilon_\theta\left(x_i, i\right)$ using a UNet, We design the Feature Interaction and Reconstruction Network to predict $\tilde{x}_r = X_{\theta}(x_s^i, i, x_v)$ during the training phase of SeDA.

Specifically, given the visual features $x_v \in \mathbb{R}^{B \times d_k}$ and the textual features $x_s \in \mathbb{R}^{B \times d_k}$, where $B$ represents the batch size and $d_k$ denotes the feature dimension. In the forward process, Gaussian noise is added step-by-step to the data $x_s$, following a Markov process with a predefined variance schedule $\{\beta_i\}_{i=1}^T$:
\begin{align}
q\left(x_s^i \mid x_s^0\right) &= \mathcal{N}\left(x_s^i ; \sqrt{\bar{\alpha}_i} x_s^0,\left(1-\bar{\alpha}_i\right) \mathbf{I}\right)
\label{eq:forward}
\end{align}
where $x_s^0 \sim q(x)$, $\mathcal{N(.)}$ means a Gaussian distribution. $\alpha_i = 1-\beta_i$ and $\bar{\alpha}_i = \prod_{j=0}^i \alpha_j$,

According to Equation \ref{eq:forward}, the noisy textual feature at step $i$ in the forward process is defined as $x_s^i=\sqrt{\bar{\alpha}_i}x_s+\sqrt{1-\bar{\alpha}_i}\epsilon$, where $\epsilon \sim \mathcal{N}(\mathbf{0},\mathbf{I})$ is the added Gaussian noise. As $i \to +\infty$, $x_s^i$ undergoes a gradual convergence towards the standard Gaussian distribution.

The fused feature representation $F_{t}$ is generated through the cross-attention mechanism, and the process can be represented as:

% First, using the cross attention mechanism, visual and textual features are effectively combined in the latent space to capture the deep relationship between visual information and textual semantics. Specifically, given the visual features $x_v$ and the textual features $x_s$, the fused feature representation $F_{t}$ is generated through the cross attention mechanism, and the process can be represented as:
\begin{equation}
   \mathbf{F}_{t} = \text{softmax}\left(\frac{Q K^T}{\sqrt{d_k}}\right) V 
\end{equation}
where ${x}_s^i$ are used as the Query(Q), and the visual features $x_v$ are used as the Key(K) and Value(V).

 $\text{SinusoidalPosEmb}(t)$ (sinusoidal position embedding function) is used to embed the time step $t$, obtaining the time feature vector $\mathbf{e}_{t} \in \mathbb{R}^{B \times d_k}$:

\begin{equation}    
\mathbf{e}_{t} = \phi(\mathbf{W}_{2} \cdot \phi(\mathbf{W}_{1} \cdot \text{SinusoidalPosEmb}(t)))
\end{equation}
where $\mathbf{W}_{1}$ and $\mathbf{W}_{2}$ are weight matrices, and $\phi$ is the activation function.

Subsequently, we concatenate the output $\mathbf{F}_{t}$ from the attention module with the encoded diffusion time feature $\mathbf{e}_{t}$ as the input to the denoising decoder:
\begin{equation}
    \mathbf{h}_{0} = [\mathbf{F}_{t}; \mathbf{e}_{t}]
\end{equation}

The denoising decoder is a multi-layer perceptron (MLP), which contains an intermediate layer with linear transformations and activation functions to encode the features, as well as a linear layer to compute the output distribution.

\subsection{Diffusion-controlled Semantic Learner (DSL)}

%由于视觉模态和文本模态之间的异构性，直接从视觉模态向文本模态迁移会有困难，因此我们提出了两阶段的对齐策略。在本模块中，通过对视觉特征中类别中心的约束，我们首先学习视觉模态中的结构信息。 
% 可以改短这句话
% Due to the heterogeneity between the visual and textual modalities, directly migrating from the visual modality to the textual modality poses challenges. To address this, we propose a two-stage alignment strategy. In this module, we first learn the structural information within the visual modality by constraining the category centers of visual features.

% 由于视觉模态和文本模态之间的异构性，直接从视觉模态向文本模态迁移会有困难，因此我们提出了异步对齐策略。在本模块中，通过对视觉特征中类别中心的约束，我们首先将视觉特征映射到语义特征空间中。

Due to the heterogeneity between the visual and textual modalities, directly mapping from the visual modality to the textual modality poses challenges. To address this, we propose a bi-stage alignment strategy. In this module, we first map the visual features space to the semantic space by constraining the category centers of the visual features.

% 与传统的扩散模型不同，我们设计Diffusive Feature Fusion and Reconstruction Network来预测反向过程中的$\hat{x}_s^0 = X_{\theta}(x_s^t,t,x_v)$。在此过程中，我们通过计算重构特征 $\hat{x}_s^0$ 和原始输入视觉特征$x_v$之间的structural consistency loss来构造表示生成 $\hat{x}_s^0$ 。
% The centroid of the visual features for category $c$ is denoted as $\mu_c$ and can be computed as:
% $$\mu_c = \frac{1}{|X_v^c|} \sum_{x \in X_v^c} x$$
% Similarly, the centroid of the reconstructed features for category $c$ is denoted as $\hat{\mu}_c$ and is computed as:
% $$\hat{\mu}_c = \frac{1}{|\hat{X}_s^c|} \sum_{\hat{x} \in \hat{X}_v^c} \hat{x}$$

% Unlike vanilla diffusion models, We design the Feature Interaction and Reconstruction Network to predict $\tilde{x}_r = X_{\theta}(x_s^i,i,x_v)$ during the training phase of ProDiffAlign.

% 根据\cite{ddim}，公式\ref{eq:mu}通过预测的\hat{x}_r^0进行计算从而得到后向过程生成的特征。\tilde{\mu}_t(x_t, x_0) = \frac{\sqrt{\bar{\alpha}_{t-1}} \beta_t}{1 - \bar{\alpha}_t} x_0 + \frac{\sqrt{\alpha_t} (1 - \bar{\alpha}_{t-1})}{1 - \bar{\alpha}_t} x_t。
% According to \cite{ddim}, Equation \ref{eq:mu} computes the features generated by the reverse process using the predicted $\hat{x}_r^0$:
% \begin{equation}
% \mu_\theta(x_r^t, \hat{x}_r^0) = \frac{\sqrt{\bar{\alpha}_{t-1}} \beta_t}{1 - \bar{\alpha}_t} \hat{x}_r^0 + \frac{\sqrt{\alpha_t} (1 - \bar{\alpha}_{t-1})}{1 - \bar{\alpha}_t} x_r^t
% \end{equation}

During this process, we construct the representation of $\tilde{x}_r$ by calculating the structural consistency loss between the interactive feature $\tilde{x}_r$ and the original input visual feature $x_v$.

The center of the visual features for category $c$ is denoted as $\mu_c$,  and similarly,  the center of the interactive features $\mu_c$ can be computed as:
\begin{equation}
\mu_c = \frac{1}{|x_v^c|} \sum_{x_v \in x_v^c} x_v
,\quad
\hat{\mu}_c = \frac{1}{|\tilde{x}_r^c|} \sum_{\tilde{x}_r \in \tilde{x}_r^c} \tilde{x}_r
\end{equation}

% The distance between the center of visual feature and the reconstructed feature can be computed by:
% \begin{equation}
% \mathcal{L}_{center} = ||\hat{\mu}_c - \mu_c||_2
% \end{equation}

% Then we compute the offset loss $\mathcal{L}_{offset}$ from the reconstructed feature to the visual feature for category $c$ by:
% \begin{equation}
% \mathcal{L}_{offset} = \sum_{x_v \in x_v^c,\hat{x}_r \in \hat{x}_r^c} ||\hat{x}_r - x_v||_1
% \end{equation}

% % 结构一致性损失为$\mathcal{L}_sc=\mathcal{L}_{center}+\mathcal{L}_{offset}$
% The structural consistency loss is given by:
% \begin{equation}
% \mathcal{L}_{SC} = \mathcal{L}_{center} + \mathcal{L}_{offset}
% \end{equation}

Next, the structural consistency loss is defined as:
\begin{equation}
    \mathcal{L}_{SC} = \|\hat{\mu}_c - \mu_c\|_2 + \sum_{x_v \in x_c^v, \tilde{x}_r \in \tilde{x}_c^r} \|\tilde{x}_r - x_v\|_1
\end{equation}
where the first term ensures the constraint of the feature centers, and the second term measures the offset between features.

Building upon cross-modal matching alignment, to enhance the model's performance in downstream visual classification tasks, we introduce a constraint cross entropy $\mathcal{L}_{CE}$ that combines visual prediction results $\hat{y}=softmax(\hat{x}_r^0)$ with real labels $y$:
\begin{equation}
     \mathcal{L}_{CE} = -\sum_{i=1}^{N} y_i \log(\hat{y}_i) \label{loss:ce}
\end{equation}

The overall structural consistency constraint loss can be defined as:
\begin{equation}
\mathcal{L}_{SCC} = \alpha_1\mathcal{L}_{SC}+\beta\mathcal{L}_{CE}
\end{equation}
where $\alpha_1$ and $\beta$ are hyperparameters used to control the contributions of the classification loss and others, respectively.

\subsection{Diffusion-controlled Semantic Translator (DST)}

% 在扩散模型的后期阶段，DCT 模块致力于在保留类别关键信息的同时学习文本特征的底层分布。DCT 模块缓解了视觉和文本模态之间的异构性。DST模态实现了从语义特征空间向文本特征空间的迁移。
In the later stage of the diffusion model, the DST module focuses on learning the underlying distribution of textual features while preserving semantic information. It facilitates the transformation from the semantic space to the textual features space.

% The diffusion model predicts the reconstructed feature $\hat{x}_r^0$ based on the input features. First we compute the Mean Squared Error(MSE) between the reconstructed feature $\hat{x}_r^0$ and the original input textual feature $x_s$. This helps to minimize the difference between the reconstructed representation and the original textual representation, ensuring that the transformed features retain important semantic information from the textual modality. 
% \begin{equation}
%     \mathcal{L}_{MSE} = ||\hat{x}_r^0 - x_s||_2^2
% \end{equation}

To optimize the underlying data generation distribution, which is typically achieved by minimizing the variational lower bound (VLB) of the negative log-likelihood, We follow the DDPM \cite{ddpm} setup and minimize the KL divergence between the two distributions $q({x}_r^i | {x}_r^{i-1}, x_r^0)$ and $p_\theta({x}_r^{i-1} | {x}_r^i)$:
\begin{equation}
    \mathcal{L}_{vlb} = D_{KL}(q({x}_r^i | {x}_r^{i-1}, x_r^0) || p_\theta({x}_r^{i-1} | {x}_r^i))
\end{equation}
To simplify the optimization process, we reformulate it into a Mean-Squared Error (MSE) loss function as follows:
\begin{equation}
    \mathcal{L}_{MSE} =\mathbb{E}_{x_s,x_s^i,x_v}[||x_s - X_{\theta}(x_s^i,i,x_v)||^2]
\end{equation}

Similar to the DSCC module, we introduce the cross-entropy constraint to assist in the training of the DCT module:
\begin{equation}
\mathcal{L}_{CT} = \alpha_2\mathcal{L}_{MSE} +\gamma\mathcal{L}_{CE}
\end{equation}
where $\alpha_2$ and $\gamma$ are hyperparameters used to control the contributions of the MSE loss and the classification loss, respectively. The $\mathcal{L}_{CE}$ has been given in Equation\ref{loss:ce}.

% 最终，这两个阶段的扩散模型过程可以表示成
Finally, the diffusion model training process for this bi-stage can be expressed as follows:
\begin{equation}
\mathcal{L} = \sum_{i=0}^{T} \begin{cases} 
\alpha_1 \mathcal{L}_{\text{SC}} + \beta \mathcal{L}_{\text{CE}}, & \text{if } i \leq t, \\ 
\alpha_2 \mathcal{L}_{\text{MSE}} + \gamma \mathcal{L}_{\text{CE}}, & \text{if } i > t.
\end{cases}
\end{equation}
where $T$ represents the diffusion model time step and $t$ represents the staged step.

% 在扩散模型推理阶段中，根据\cite{ddim}，公式\ref{eq:mu}中对$\mu_{\theta}$的计算可以通过FIRN预测特征$\tilde{x}_r^0$来计算：\begin{equation}
% \mu_\theta(x_r^i, \tilde{x}_r^0) = \frac{\sqrt{\bar{\alpha}_{i-1}} \beta_t}{1 - \bar{\alpha}_i} \tilde{x}_r^0 + \frac{\sqrt{\alpha_i} (1 - \bar{\alpha}_{i-1})}{1 - \bar{\alpha}_i} x_r^i
% \end{equation}.通过a step-by-step reverse denoising operation\[
% \hat{x}_r^T \xrightarrow{p_\theta(\hat{x}_r^{T-1} | \hat{x}_r^T)} \hat{x}_r^{T-1} \xrightarrow{p_\theta(\hat{x}_r^{T-2} | \hat{x}_r^{T-1})} \hat{x}x_r^{T-2} \xrightarrow{\dots} \hat{x}_r^1 \xrightarrow{p_\theta(\hat{x}_r^0 | \hat{x}_r^1)} \hat{x}_r^0.
% \]最终得到对齐后的特征\hat{x}_r^0。

\subsection{Inference phase of SeDA}

In the inference process, the goal is to iteratively reconstruct the original data by optimizing the likelihood $p_\theta(x_r^0)$. The reverse process is defined by:
\begin{align}
p_\theta\left(x_r^{i-1} \mid x_r^i\right) &= \mathcal{N}\left(x_r^{i-1} ; \mu_\theta\left(x_r^i, \tilde{x}_r\right), \tilde{\beta}_i \mathbf{I} \right)
\end{align}
where $\mu_{\theta}$ is the predicted mean, and $\tilde{\beta}_i = \frac{1 - \bar{\alpha}_{i-1}}{1 - \bar{\alpha}_i} \beta_i$ is the variance term. 

According to \cite{ddim}, $\mu_{\theta}$ can be calculated using the predicted feature $\tilde{x}_r$ from the FIRN module:
\begin{equation}
    \mu_\theta(x_r^i, \tilde{x}_r) = \frac{\sqrt{\bar{\alpha}_{i-1}} \beta_t}{1 - \bar{\alpha}_i} \tilde{x}_r + \frac{\sqrt{\alpha_i} (1 - \bar{\alpha}_{i-1})}{1 - \bar{\alpha}_i} x_r^i
\end{equation}

By performing a step-by-step reverse denoising operation:
\begin{equation}
    \hat{x}_r^T \xrightarrow{p_\theta(\hat{x}_r^{T-1} | \hat{x}_r^T)} \hat{x}_r^{T-1} \xrightarrow{\dots} \hat{x}_r^1 \xrightarrow{p_\theta(\hat{x}_r^0 | \hat{x}_r^1)} \hat{x}_r^0
\end{equation}
the aligned feature $\hat{x}_r^0$ is finally obtained.

Subsequently, the aligned feature $\hat{x}_r^0$ is fed into a fully connected classifier to compute the predicted logits for the final classification.

% -------------性能大表----------------
\begin{table*}[htbp]
\belowrulesep=0pt
\aboverulesep=0pt
\centering 
% 减少空行
% \vspace{-0.2 cm}
\renewcommand{\arraystretch}{1.1}
\setlength{\tabcolsep}{2.9mm}
{
\begin{tabular}{c|c|clcl|cccc|clcl}
    \toprule
    \multirow{2}{*}{Method}               & \multicolumn{1}{c|}{\multirow{2}{*}{Model}}
    % \multirow{2}{*}{\textbf{Method}} 
    % & \multirow{2}{*}{\textbf{Model}} 
        & \multicolumn{4}{c|}{\textbf{VIREO Food-172}} 
        & \multicolumn{4}{c|}{\textbf{NUS-WIDE}} 
        &\multicolumn{4}{c}{\textbf{MSRVTT}} 
        \\ \cline{3-14}        
        && 
        \multicolumn{2}{c}{\textbf{Acc-1}} & \multicolumn{2}{c|}{\textbf{Acc-5}} & \multicolumn{1}{c}{\textbf{Pre-1}} & \multicolumn{1}{c}{\textbf{Pre-5}} & \multicolumn{1}{c}{\textbf{Rec-1}} & \multicolumn{1}{c|}{\textbf{Rec-5}} &
        \multicolumn{2}{c}{\textbf{Acc-1}} & \multicolumn{2}{c}{\textbf{Acc-5}}  \\ 
        \midrule
        \multirow{5}{*}{\begin{tabular}[c]{@{}c@{}}Visual\\ Modal\\ Backbone
\end{tabular}} 
        & ResNet-50 (CVPR'16)                       & \multicolumn{2}{c}{81.58}          & \multicolumn{2}{c|}{95.02}          & \multicolumn{1}{c}{78.56}          & \multicolumn{1}{c}{39.12}          & \multicolumn{1}{c}{44.04}             & 86.42  &
        \multicolumn{2}{c}{51.37}          & \multicolumn{2}{c}{79.03}  \\   
        
        % & WRN (BMVC'16)                            & \multicolumn{2}{c}{82.29}          & \multicolumn{2}{c|}{95.52}          & \multicolumn{1}{c}{78.73}          & \multicolumn{1}{c}{39.36}          & \multicolumn{1}{c}{44.01}             & 85.33   &
        % \multicolumn{2}{c}{49.57}          & \multicolumn{2}{c}{79.93}          \\   
        % & WISeR (MMM'18)                           & \multicolumn{2}{c}{82.76}          & \multicolumn{2}{c|}{96.51} & \multicolumn{1}{c}{78.90}          & \multicolumn{1}{c}{39.45}          & \multicolumn{1}{c}{44.14}             & 85.52    &
        % \multicolumn{2}{c}{50.80}          & \multicolumn{2}{c}{81.60}     \\     
        & RepVGG (CVPR'21)                             & \multicolumn{2}{c}{83.47}          & \multicolumn{2}{c|}{96.03}          & \multicolumn{1}{c}{79.71}          & \multicolumn{1}{c}{39.44}          & \multicolumn{1}{c}{44.82}             & 85.58    &
        \multicolumn{2}{c}{50.96}          & \multicolumn{2}{c}{77.12}         \\  
        & RepMLPNet (CVPR'22)                             & \multicolumn{2}{c}{83.36}          & \multicolumn{2}{c|}{96.22}          & \multicolumn{1}{c}{80.10}          & \multicolumn{1}{c}{40.53}          & \multicolumn{1}{c}{44.82}             & 87.69    &
        \multicolumn{2}{c}{51.07}          & \multicolumn{2}{c}{77.16}         \\

        & ViT-B/16 (ICLR'20)                           & \multicolumn{2}{c}{85.37}          & \multicolumn{2}{c|}{97.29}          & \multicolumn{1}{c}{80.46}          & \multicolumn{1}{c}{40.57}          & \multicolumn{1}{c}{45.50}             & 87.96    &
        \multicolumn{2}{c}{53.25}          & \multicolumn{2}{c}{81.85}      \\ 
        & VanillaNet (NIPS'24)                           & \multicolumn{2}{c}{84.51}          & \multicolumn{2}{c|}{96.04}          & \multicolumn{1}{c}{80.12}          & \multicolumn{1}{c}{39.46}          & \multicolumn{1}{c}{45.65}             & 85.73      &
        \multicolumn{2}{c}{52.47}          & \multicolumn{2}{c}{80.02}   
        \\
        \midrule

        \multirow{11}{*}{\begin{tabular}[c]{@{}c@{}}Alignment\\Framework\end{tabular}}                   & ATNet (MM'19)                & \multicolumn{2}{c}{85.67}          & \multicolumn{2}{c|}{96.81}          & \multicolumn{1}{c}{80.78}          & \multicolumn{1}{c}{39.89}          & \multicolumn{1}{c}{45.59}             & 86.55    &
        \multicolumn{2}{c}{54.45}          & \multicolumn{2}{c}{82.68}         
        \\                         

        & CLIP (PMLR'21)                
        & \multicolumn{2}{c}{85.56} & \multicolumn{2}{c|}{96.98}          & \multicolumn{1}{c}{81.64} & \multicolumn{1}{c}{40.87} & \multicolumn{1}{c}{46.25}    & 88.78 &
        \multicolumn{2}{c}{52.89}          & \multicolumn{2}{c}{81.66}  \\  
        
        % & SWD (CVPR'19)                
        % & \multicolumn{2}{c}{87.60} & \multicolumn{2}{c|}{97.90}          & \multicolumn{1}{c}{82.52} & \multicolumn{1}{c}{41.02} & \multicolumn{1}{c}{47.04}    & 88.97  &
        % \multicolumn{2}{c}{55.55}          & \multicolumn{2}{c}{83.81} \\                                  
        % & SSAN (MM'20)               
        % & \multicolumn{2}{c}{87.09} & \multicolumn{2}{c|}{97.73}          & \multicolumn{1}{c}{81.62} & \multicolumn{1}{c}{40.55} & \multicolumn{1}{c}{46.25}    & 88.02  &
        % \multicolumn{2}{c}{54.18}          & \multicolumn{2}{c}{83.55} \\                                  
        % & CDD (CVPR'19)               
        % & \multicolumn{2}{c}{86.01} & \multicolumn{2}{c|}{96.97}          & \multicolumn{1}{c}{81.60} & \multicolumn{1}{c}{40.52} & \multicolumn{1}{c}{46.25}    & 87.96  &
        % \multicolumn{2}{c}{54.68}          & \multicolumn{2}{c}{81.84} \\                                  
   
        & TEAM (MM'22)               
        & \multicolumn{2}{c}{87.70} & \multicolumn{2}{c|}{97.85}      & \multicolumn{1}{c}{81.98} & \multicolumn{1}{c}{40.80} & \multicolumn{1}{c}{46.48}    & 88.54  &
        \multicolumn{2}{c}{55.08}          & \multicolumn{2}{c}{84.12} \\ 

        & ITA (CVPR'22)               
        & \multicolumn{2}{c}{87.82} & \multicolumn{2}{c|}{97.89}          & \multicolumn{1}{c}{82.65} & \multicolumn{1}{c}{41.40} & \multicolumn{1}{c}{46.99}    & 89.13  &
        \multicolumn{2}{c}{55.05}          & \multicolumn{2}{c}{84.05} \\ 

        & SDM (CVPR'23)               
        & \multicolumn{2}{c}{87.63} & \multicolumn{2}{c|}{97.78}          & \multicolumn{1}{c}{82.64} & \multicolumn{1}{c}{41.20}& \multicolumn{1}{c}{47.03}    & 89.31 &
        \multicolumn{2}{c}{54.82}          & \multicolumn{2}{c}{\textbf{84.28}}  \\  

        & MM-PKD (CVPR'23)              
        & \multicolumn{2}{c}{87.89} & \multicolumn{2}{c|}{96.96}          & \multicolumn{1}{c}{81.76} & \multicolumn{1}{c}{41.31} & \multicolumn{1}{c}{47.24}    & 89.11  &
        \multicolumn{2}{c}{54.77}          & \multicolumn{2}{c}{82.45} \\

        & C2KD (CVPR'24)               
        & \multicolumn{2}{c}{87.83} & \multicolumn{2}{c|}{98.06}          & \multicolumn{1}{c}{82.77} & \multicolumn{1}{c}{41.25} & \multicolumn{1}{c}{47.20}    & 89.35  &
        \multicolumn{2}{c}{55.15}          & \multicolumn{2}{c}{82.21} \\   

        & MGCC (AAAI'24)               
        & \multicolumn{2}{c}{87.80} & \multicolumn{2}{c|}{97.87}          & \multicolumn{1}{c}{82.09} & \multicolumn{1}{c}{41.05} & \multicolumn{1}{c}{46.69}    & 89.14  &
        \multicolumn{2}{c}{54.88}          & \multicolumn{2}{c}{83.67} \\  

        & MoMKE (MM'24)               
        & \multicolumn{2}{c}{87.97} & \multicolumn{2}{c|}{97.09}          & \multicolumn{1}{c}{81.72} & \multicolumn{1}{c}{41.11} & \multicolumn{1}{c}{46.50}    & 89.14  &
        \multicolumn{2}{c}{54.57}          & \multicolumn{2}{c}{82.72} \\  
        
        \cline{2-14}     
        
        & SeDA$_{RN50}$               
        & \multicolumn{2}{c}{86.01} & \multicolumn{2}{c|}{96.97}  & \multicolumn{1}{c}{81.60} & \multicolumn{1}{c}{40.52} & \multicolumn{1}{c}{46.25} & 87.96  &
        \multicolumn{2}{c}{54.68}          & \multicolumn{2}{c}{81.84}   \\

        & SeDA$_{ViT16}$                & 
        \multicolumn{2}{c}{\textbf{89.19}} & 
        \multicolumn{2}{c|}{\textbf{98.07}}          & \multicolumn{1}{c}{\textbf{83.46}} & 
        \multicolumn{1}{c}{\textbf{41.60}} &
        \multicolumn{1}{c}{\textbf{47.72}}    & 
        \textbf{90.21}   &
        \multicolumn{2}{c}{\textbf{57.09}}          & \multicolumn{2}{c}{83.91}          \\ 
        \bottomrule
\end{tabular}}
\caption{Performance comparison of algorithms on VIREO Food-172, NUS-WIDE and MSRVTT datasets. Metrics are Top-1/Top-5 Accuracy (Acc), Precision (Pre), and Recall (Rec). The best performance of each indicator has been highlighted in bold} 
% \vspace{-0.3cm}
\label{table:performance}
\end{table*}

\section{Experiment}
\subsection{Experiment Setting}

\subsubsection{Datasets}
% 为了验证跨模态扩散对齐方法的有效性和通用性，我们将算法应用于图像分类任务上，在三个跨模态的数据集上进行实验。详情如下。
To assess the effectiveness and generality of SeDA, we conducted experiments on image and video classification tasks across three datasets. Details are provided below.
\begin{itemize}
\item \textbf{VIREO Food-172\cite{vireo172}:} A single-label dataset with 110,241 food images in 172 categories and an average of three text descriptions per image. It includes 66,071 training and 33,154 test images.

\item \textbf{NUS-WIDE\cite{Nus-wide}:} A multi-label dataset of 203,598 images (after filtering) in 81 categories, with textual tags from a 1000-word vocabulary. It has 121,962 training and 81,636 test images.

\item \textbf{MSRVTT\cite{msrvtt}:} A video dataset with 10,000 YouTube clips and 200,000 captions. We used 7,010 videos for training and 2,990 for testing.
\end{itemize}

\subsubsection{Evaluation Metric}
%要根据我们采用的评价指标改  
% 参考AAAI25 可以改短

% 在单标签分类任务中，本文实验采用准确率作为性能评估指标，具体计算公式如下：

% 其中，TP是模型预测为真的正例样本的数量，FN是模型预测为假的反例样本的数量，TN是模型预测为真的反例样本的数量，FP是模型预测为假的正例样本的数量。
% 在多标签分类任务中，本文实验采用精准率和召回率作为性能评估指标，具体计算公式如下：
% 对于上述两个指标，本文分别计算TOP-1和TOP-5的平均值作为实验结果展示。

% 对于Vireo-Food172和MSRVTT单分类预测任务上的数据集，我们根据之前的工作使用top1和-5的精度\cite{}来评估性能。对于NUS-WIDE数据集，我们参照\cite{}中，计算Top-1和-5的overall精度和召回率。

% In the single-label classification task, we use accuracy as the performance evaluation metric. The specific calculation formula is as follows:
% \begin{equation}
% Accuracy = \frac{TP+TN}{TP+TN+FP+FN}
% \end{equation}
% where, $TP$ is the number of true positive samples predicted by the model, $FN$ is the number of false negative samples predicted by the model, $TN$ is the number of true negative samples predicted by the model, and $FP$ is the number of false positive samples predicted by the model.
% In the multi-label classification task, we use precision and recall as performance evaluation metrics, following the methods described in \cite{eval_1,eval_2}. The specific calculation formulas are as follows:
% \begin{equation}
% Precision = \frac{TP}{TP+FP}
% \end{equation}
% \begin{equation}
% Recall = \frac{TP}{TP+FN}
% \end{equation}
% For these metrics mentioned above, we calculate the average of TOP-1 and TOP-5 values to present the experimental results.

For the VIREO Food-172 and MSRVTT datasets of the single-class prediction task, we use the accuracy of Top-1 and -5 following \cite{ATNet}. For the multi-label dataset NUS-WIDE, we calculate Top-1 and -5 overall precision and recall following \cite{eval_1,eval_2}.

% \subsubsection{Implementation Details}

% % 在Visual Modal Backbone的实现，均参照对应论文的方法实现（）。The implementation of Alignment Framework employs ViT-B/16[] for encoding the visual channel and BERT[] for encoding the semantic channel.分类器为一个具有单层全连接的网络。

% In the implementation of the Visual Modal Backbone, methods are based on the approaches described in the corresponding papers (ResNet-50 \cite{resnet18},  WRN \cite{wrn}, WISeR \cite{WISeR}, RepVGG \cite{repvgg}, RepMLPNet \cite{repmlpnet}, ViT-B/16 \cite{vit}, VanillaNet \cite{vanillanet}). The implementation of the Alignment Framework uses ViT-B/16 \cite{vit} for encoding the visual channel and BERT \cite{bert} for encoding the semantic channel. We adapt the cross-modal alignment method from the original to our LUPI task (ATNet \cite{ATNet}, CLIP \cite{CLIP}, SWD \cite{SWD}, SSAN \cite{SSAN}, CDD \cite{cdd}, SDM\cite{sdm}, TEAM \cite{TEAM}, ITA \cite{ita}, MM-PKD \cite{MM-PKD}, C2KD \cite{C2KD}, MGCC \cite{MGCC}, MoMKE \cite{MoMKE}). The classifier is a single-layer fully connected network.

\subsubsection{Implementation Details}
% 在本实验中，我们选择Adam作为模型的优化器，其中Adam的权重衰减设置为1e-4，所有神经网络的学习率设置为1e-4到5e-5。每四轮训练，学习率就会衰减到原来的0.5倍。对于训练策略中提到的损失的权重，我们在0.1到2.0之间选择$\alpha_{sim}$和$\alpha_{align}$，$\beta$和$\gamma$的值在[中选择0.1、0.2、0.5、1.0]。我们在 NVIDIA Tesla V100 上进行实验，小批量输入 64 张图像，每个阶段训练 40 轮。对于 Vireo-Food172，基于 vit 的模型需要 $ \sim $8 小时来训练。对于多分类数据集NUS-WIDE，我们将BCE损失的正样本损失权重设置为20到150，基于vit的模型需要$ \sim $12小时来训练。对于视频数据集MSRVTT，需要$ \sim $12小时来训练。

In this experiment, we chose Adam as the optimizer for the model, with a weight decay of 1e-4. The learning rate for all neural networks was set between 1e-4 and 5e-5. The learning rate decayed to half of its original value every four training epochs. For the loss weights mentioned in the training strategy, we selected $\alpha_{1}$ and $\alpha_{2}$ between 0.1 and 2.0, the time step $T$ between 900 and 1500, the staged step $t$ between 0 and 500, while $\beta$ and $\gamma$ were chosen from [0.5, 1.0, 1.5, 2.0]. Our experiments were conducted on a single NVIDIA Tesla V100 GPU, using PyTorch 1.10.2, and the batch size is 64.

% For Vireo-Food172, a model based on ViT took approximately $ \sim $8 hours to train. For the multi-class dataset NUS-WIDE, we set the positive sample weight of the BCE loss between 20 and 150, and a ViT-based model took approximately $ \sim $12 hours to train. For the video dataset MSRVTT, it also took approximately $ \sim $12 hours to train.

\subsection{Performance Comparison}

% 我们对比了9种visual modal backbone和12种对齐方法。
% 这个记得改
% We compared 7 visual modality backbones and 12 alignment methods. The following results can be observed from Table\ref{table:performance}. 

We conducted a comprehensive comparison involving 5 visual modal backbones and 9 alignment frameworks, with results summarized in Table \ref{table:performance}. The visual modal backbones were implemented based on the methods outlined in their respective papers, including ResNet-50 \cite{resnet18}, RepVGG \cite{repvgg}, RepMLPNet \cite{repmlpnet}, ViT-B/16 \cite{vit}, and VanillaNet \cite{vanillanet}. For the alignment framework, ViT-B/16 was used to encode the visual channel, while BERT \cite{bert} was employed for the textual channel. In addition, we validated the effectiveness of our method on ResNet-50. The cross-modal alignment methods were adapted from their original designs to fit the LUPI task, including ATNet \cite{ATNet}, CLIP \cite{CLIP}, TEAM \cite{TEAM}, ITA \cite{ita}, SDM \cite{sdm}, MM-PKD \cite{MM-PKD}, C2KD \cite{C2KD}, MGCC \cite{MGCC}, and MoMKE \cite{MoMKE}. A fully connected single-layer network served as the classifier.

% FedFSA是一个通用框架，它可以结合各种基于知识蒸馏的方法，如FedAvg和FedETF，为它们带来性能提高，这展示了其与模型无关的能力
% 基于模型校准的方法通常优于基于知识蒸馏的方法，正如FedCSPC，CLIP2FL和FedFSA所证明的那样。这是因为它们都在努力利用来自多个来源的信息来训练一个广义模型。•FedETF采用统一的单纯形等角紧框架分类器，通常比基于数据驱动知识的方法（FedProc，FedNTD，MOON）能得到更好的结果。这可能是由于它们避免了由数据差异和模型本身的固有限制所造成的知识质量差的问题。•随着数据源数量的增加，性能往往会下降。这是由于不同数据分布之间的差异被放大了。FedFSA在性能上保持了其优势，充分证明了其校准机制的有效性。

% 基于不同的骨干网，VSCNet 取得了显著的改进。它展示了修改细粒度关联对图像分类的影响，并显示了 VSCNet 的骨干无关性。 - 在两个数据集中，VSCNet 的性能普遍优于其他算法。这是合理的，因为 VSCNet 能够生成 “视觉-语义-类别 ”层次结构，利用预测的视觉-语义关联来准确推断图像类别。 - 与其他视觉骨干网相比，ViT 和 VanillaNet 的性能有了显著提高。 得益于变换器和激活函数的判别能力，ViT 和 VanillaNet 在两个数据集上的性能都比其他方法高出 8%。 - 显式配准方法通常比隐式配准方法获得更好的性能。这表明，显式特征配准比隐式正则化更有利于视觉特征的学习。 - 细粒度配准方法、区域定位和信息聚合都会影响其性能。与 VSCNet 相比，MGCC 显示出较高的前 5 名性能，但较差的信息聚合降低了前 1 名的性能。

\begin{itemize}
% 我们所提出的方法ProDiffAlign在三个数据集上的性能普遍优于其他算法，这得益于我们的方法能够逐步学习到文本特征的底层分布，缓解类别之间的语义混淆。
    \item \textbf{The proposed method, SeDA, outperforms other algorithms across three datasets.} This is attributed to our method's ability to learn the underlying distribution of textual features, helping alleviate semantic confusion between categories.

% % visual backbone中 vit表现好
%     \item \textbf{ViT and VanillaNet show significantly improved performance compared to other visual backbones.} Taking advantage of the transformer’s discriminative power and activation function, ViT and VanillaNet outperformed other methods in VIREO Food-172 up to 5\%.

% 模型无关
    \item \textbf{SeDA is a general framework that can combine various visual backbones,} such as ViT-B/16 and ResNet-50, to bring them performance gains, which showcases its model-agnostic capability.

% Alignment Framework中的方法普遍比Visual Modal Backbone表现好，这是由于更具判别力的文本信息的加入改善了视觉表征。
    \item \textbf{Models in the Alignment Framework generally outperform the Visual Modal Backbone.} This can be attributed to the incorporation of more discriminative textual information, which effectively optimizes the visual representation.

% 我们的方法在Top-1准确率上的提升明显优于Top-5，主要因为其聚焦于文本分布的学习，致力于准确提供最具判别力的第一个标签，而非专注于提升检索排名中的精度。
    \item \textbf{SeDA shows a more significant improvement in Top-1 accuracy than in Top-5.} This is mainly because it focuses on learning the distribution of textual features, aiming to accurately provide the most discriminative first label rather than focusing on improving the precision in retrieval ranking.

% % 在基于特征差异度量的方法中，基于对比学习的方法（ATNet,SWD,SSAN,CDD）普遍优于基于距离度量的特征分布的方法（CLIP,SDM,TEAM,ITA）。这可能是由于对比学习方法通过构建正负样本对在捕获跨模态语义相似性和应对复杂数据分布上更具优势。
%     \item \textbf{In feature discrepancy measurement methods, contrastive learning-based methods (CLIP, SDM, TEAM, ITA) generally outperform distance metric-based feature distribution methods (ATNet, SWD, SSAN, CDD).} This is likely because contrastive learning methods, by constructing positive and negative sample pairs, are more effective at capturing cross-modal semantic similarity and handling complex data distributions.
% % 基于知识蒸馏的方法（C2KD,MM-PKD）在vireo-172数据集上表现出色，但它们对对齐质量和融合技术的依赖可能会限制高度多样化数据集的可扩展性。
%     \item Knowledge distillation-based methods (C2KD, MM-PKD) perform well on the Vireo-172 dataset, showcasing their ability to effectively transfer knowledge between modalities through teacher-student frameworks.  \textbf{But their reliance on alignment quality and fusion techniques may limit scalability in highly diverse datasets.}

\end{itemize}

\subsection{Ablation Study}

\begin{table}[t]
\centering 
% \vspace{-0.2 cm}
\setlength{\tabcolsep}{1.2mm}
\renewcommand{\arraystretch}{0.9}
{{\begin{tabular}{ccccccc}
\toprule
\multirow{2}{*}{\textbf{Method}} & \multicolumn{2}{c}{\textbf{VIREO Food-172}} & \multicolumn{4}{c}{\textbf{NUS-WIDE}}    
\\ \cmidrule(l){2-7}  
& \multicolumn{1}{c}{\textbf{{Acc-1}}} & \textbf{{Acc-5}} &
\multicolumn{1}{c}{\textbf{{Pre-1}}} & \multicolumn{1}{c}{\textbf{{Pre-5}}} & \multicolumn{1}{c}{\textbf{{Rec-1}}} & \textbf{{Rec-5}}
\\ \midrule
Base  & \multicolumn{1}{c}{85.37}   & 97.29 & \multicolumn{1}{c}{80.46} & \multicolumn{1}{c}{40.57} & \multicolumn{1}{c}{45.50} & 87.96 
\\ 
+T  & \multicolumn{1}{c}{88.70}     & 96.93  
& \multicolumn{1}{c}{82.63} & \multicolumn{1}{c}{41.23} & \multicolumn{1}{c}{47.19} & 89.37 
\\ 
+T+I  & \multicolumn{1}{c}{89.12} & 97.01 
& \multicolumn{1}{c}{82.74} & \multicolumn{1}{c}{41.20} & \multicolumn{1}{c}{47.26} & 89.48 
\\ 
+T+I+L  & \multicolumn{1}{c}{\textbf{89.19}} & \textbf{98.07}       & \multicolumn{1}{c}{\textbf{83.46}} & \multicolumn{1}{c}{\textbf{41.60}} & \multicolumn{1}{c}{\textbf{47.72}} & \textbf{90.21}  
\\ \bottomrule
\end{tabular}}}
\caption{Results of ablation study. The evaluation indexes are the same as those in Table 1. The best performance is marked in bold.}
\label{t3}
\end{table}

This section examines the performance of different modules based on ViT-B/16 in SeDA, with the results presented in Table \ref{t3}.

\begin{itemize}
\item \textbf{On the VIREO Food-172 dataset, Diffusion-controlled Semantic Translator (+T) module played a key role in improving Acc-1, achieving a 3.7\% increase compared to the baseline method.} However, there was a slight decrease in Acc-5.

\item \textbf{Adding the Progressive Feature Interaction Network (+I) module led to improvements across both datasets.} This indicates that it (+I) further enhances feature interaction capabilities.

\item \textbf{Diffusion-Controlled Semantic Learner (+L) module ensured a certain improvement in the Top-1 metric while achieving a more significant enhancement in the Top-5 metric.} Notably, on the NUS-WIDE dataset, Rec-5 improved by 2.6\% compared to the baseline method.
\end{itemize}

\subsection{In-depth Analysis}

\subsubsection{Robustness of SeDA on Hyperparameters}

% 本节评估FedFSA在不同超参数下的鲁棒性。我们从{1、2、3}、{0.1、0.5、1、5}、{0.01、0.1、1}和{0、0.3、0.5、0.7}中选择Nυ、权重参数α、η和κ。如图3所示，DMANet在很大范围内对超参数变化不敏感，表明其在超参数选择方面具有很强的鲁棒性。对于$\beta$，该模型在$\beta=1.5$时表现最好。这是因为较低$\beta$值可能导致对某些特定特征的过度依赖而忽略了类别信息，而较高的$\beta$值可能导致过于关注类别信息而破坏对特征分布空间的学习。此外，该模型在扩散模型时间步长设置为1500时表现最好，因为较小的时间步长无法充分学习到有用的信息，实现视觉模态像文本模态的迁移，较大的时间步长可能引入其他噪声。值得注意的是，融合两阶段学习是有益的，但第二阶段时间步长过长可能会导致原始特征的结构信息被破坏，使得性能有一定的下降。

This section evaluates the robustness of SeDA in different hyperparameters. We select the weight parameter $\gamma$, time step $T$ and stages step $t$ from \{0.5,1.0,1.5,2.0\}, \{900,1200,1500,1800\} and \{0,300,500\}. \textbf{SeDA is largely insensitive to changes in hyperparameters, demonstrating strong robustness in hyperparameter selection.} For $\gamma$, the model performs best when $\gamma=1.5$ 
This is because lower $\gamma$ values tend to rely overly on specific features at the expense of class-level information, whereas higher $\gamma$ values place excessive emphasis on class information, disrupting the learning of the feature distribution space. Furthermore, the model achieves its best performance when the diffusion model time step $T$ is set to 1500, as smaller $T$ fails to fully learn useful information for transferring the visual modality to the text modality, while larger $T$ may introduce additional noise. Notably, \textbf{two-stage learning is beneficial, but excessively long staged step $t$ in the second stage may damage the structural information of the original features, leading to a slight drop in performance.}

\begin{figure}[t]
\begin{center}
% [width=0.5\textwidth,height=0.5\textwidth]
\includegraphics[width=0.48\textwidth]{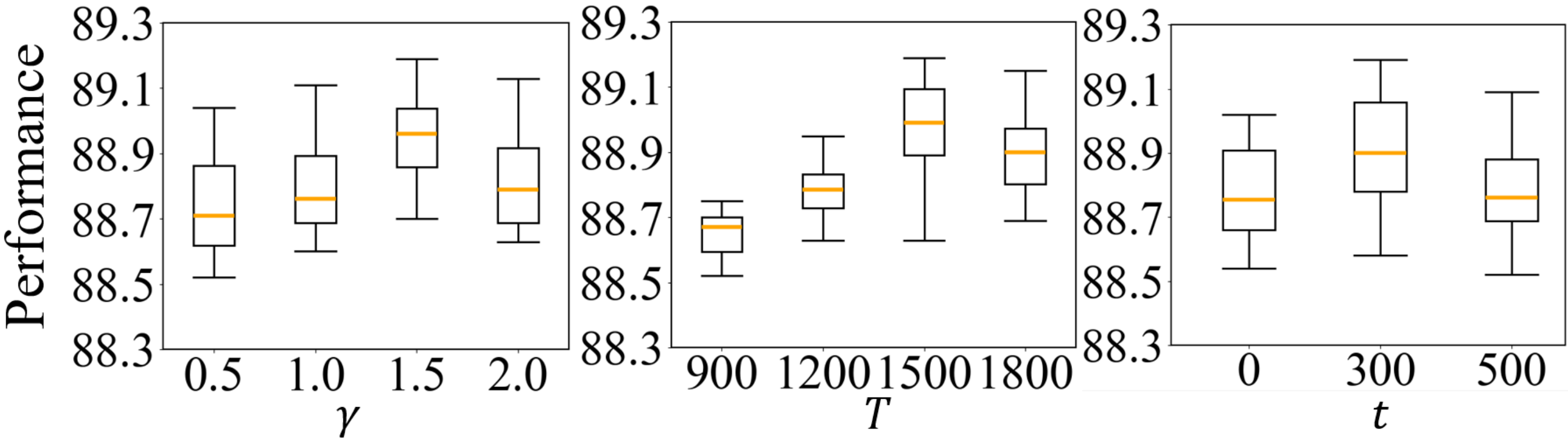}
\end{center}
% \vspace{-0.5cm}
\caption{The impact of hyperparameters on performance. The weight parameter $\gamma$, the time step $T$ and the staged step $t$ are turned from \{0.5,1.0,1.5,2.0\}, \{900,1200,1500,1800\}, \{0,300,500\} on VIREO Food-172, respectively.}
% \vspace{-0.5cm}
\label{fig:hyperparameters}
\end{figure}

% -----------不同模块不同方法-----------
% 这个表要不要加？名字需要改，文字描述是否太长？
\begin{table*}[h]
\centering 
% \vspace{-0.2 cm}

\renewcommand{\arraystretch}{0.9}
\begin{tabular}{>{\centering\arraybackslash}p{4cm}|>{\centering\arraybackslash}p{10cm}|>{\centering\arraybackslash}p{1cm}|>{\centering\arraybackslash}p{1cm}}
\hline
\textbf{Strategy} &\textbf{Method} & \textbf{Acc-1} & \textbf{Acc-5} \\
\hline
\textbf{Base} & ViT-B/16 & 85.37 & 97.29 \\ \hline
\multirow{3}{*}{\textbf{\begin{tabular}[c]{@{}c@{}}Different\\Training Strategies\end{tabular}}}
 & Freezing the visual and the textual backbone model
 & 87.27 & 93.45 \\
& Training the visual backbone, freezing the textual backbone & 88.67 & 94.84 \\
& Training the visual and the textual backbone model & 88.70 & 96.93 \\
\hline
\multirow{2}{*}{\textbf{\begin{tabular}[c]{@{}c@{}}Different Diffusion Model\\Network Structures\end{tabular}}}  & 
Using UNet as the diffusion model network  & 86.98 & 92.22 \\
& Using MLP as the diffusion model network & 87.27 & 93.45 \\
\hline
\multirow{4}{*}{\textbf{\begin{tabular}[c]{@{}c@{}}Different Modal\\Interaction Methods\end{tabular}}} & 
Using Concat for modal interaction & 88.70 & 96.93 \\ 
&Using Self Attention for modal interaction & 88.21 & 96.46 \\
& Using Multimodal Transformer for modal interaction & 87.62 & 95.44 \\
& Using Cross Attention for modal interaction & 89.12 & 97.01 \\
\hline
\multirow{3}{*}{\textbf{\begin{tabular}[c]{@{}c@{}}Different Stage\\Training Methods\end{tabular}}} & 
Using a single-stage method with only the DSL module & 88.97 & \textbf{98.13} \\
& Using a single-stage method with only the DST module  & 89.12 & 97.01 \\
% 这一行可以删掉
% & Bi-staged method with DST module first and DSE module second. & 89.07 & 98.05 \\
& Bi-stage method with DSL module first and DST module second. & \textbf{89.19} & 98.07 \\
\hline
\end{tabular}
\caption{Results from the combined experiment using different strategies and constraint functions on the VIREO Food-172 dataset.}
\vspace{-0.3 cm}
\label{tab:different_method}
\end{table*}

\subsubsection{The Effect of different methods in each module}

To further investigate the impact of different methods in the DSL, DST, and PFIN modules on model performance, we used ViT-B/16 as the baseline and conducted experiments on the VIREO Food-172 dataset, as shown in Tabel \ref{tab:different_method}. Four aspects were analyzed: training strategies, diffusion model network structures, modal interaction methods, and stage training methods. Fixing text features limits flexibility, while joint optimization improves cross-modal understanding. UNet is better suited for pixel-level learning, while MLP, which outperforms UNet, is more effective for feature-level optimization. Cross Attention proved superior for modal interaction, effectively capturing correlations. Bi-stage framework effectively integrates semantic information from visual and text features, enhancing alignment and classification capabilities.

\subsection{Case Study}

\subsubsection{Performance of Visual to Textual Feature Distribution Transformation} 

\begin{figure}[t]
\begin{center}
% [width=0.5\textwidth,height=0.5\textwidth]
\includegraphics[width=0.48\textwidth]{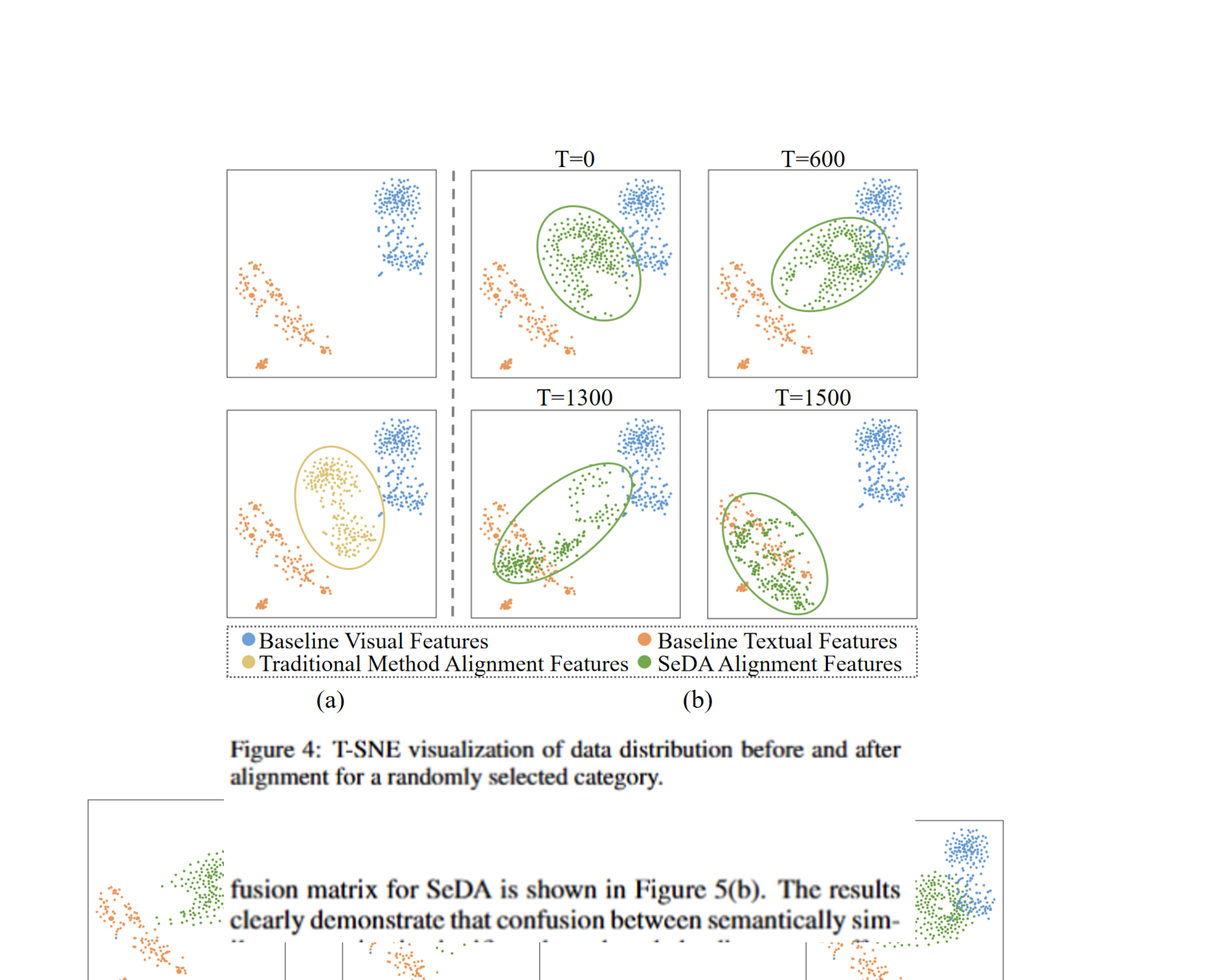}
\end{center}
% \vspace{-0.5cm}
\caption{T-SNE visualization of data distribution before and after alignment for a randomly selected category.}
% \vspace{-0.5cm}
\label{fig:class}
\end{figure}

We randomly selected a category to compare the textual and visual features of the ViT-B/16, features aligned using traditional L2-norm alignment (as shown in Figure \ref{fig:class}(a)), and features reconstructed by our method during the diffusion model sampling process across different time steps (Figure \ref{fig:class} (b)). The feature distribution produced by traditional alignment methods demonstrates minimal changes, with significant discrepancies still evident between the aligned visual and textual features. This suggests that traditional methods fail to bridge the gap between the two modalities fully. In contrast, the features reconstructed using our method initially capture the semantic characteristics of visual features, gradually moving toward textual features as the sampling process progresses. This dynamic migration process ensures that visual features are effectively transformed into a semantic space consistent with textual features. Furthermore, our method not only achieves accurate mapping between the two modalities but also enhances the expressiveness of the textual features by preserving and utilizing semantic information. 

% 跨模态迁移能力

% 语义混淆分析
\begin{figure}[t]
\begin{center}
% [width=0.5\textwidth,height=0.5\textwidth]
\includegraphics[width=0.48\textwidth]{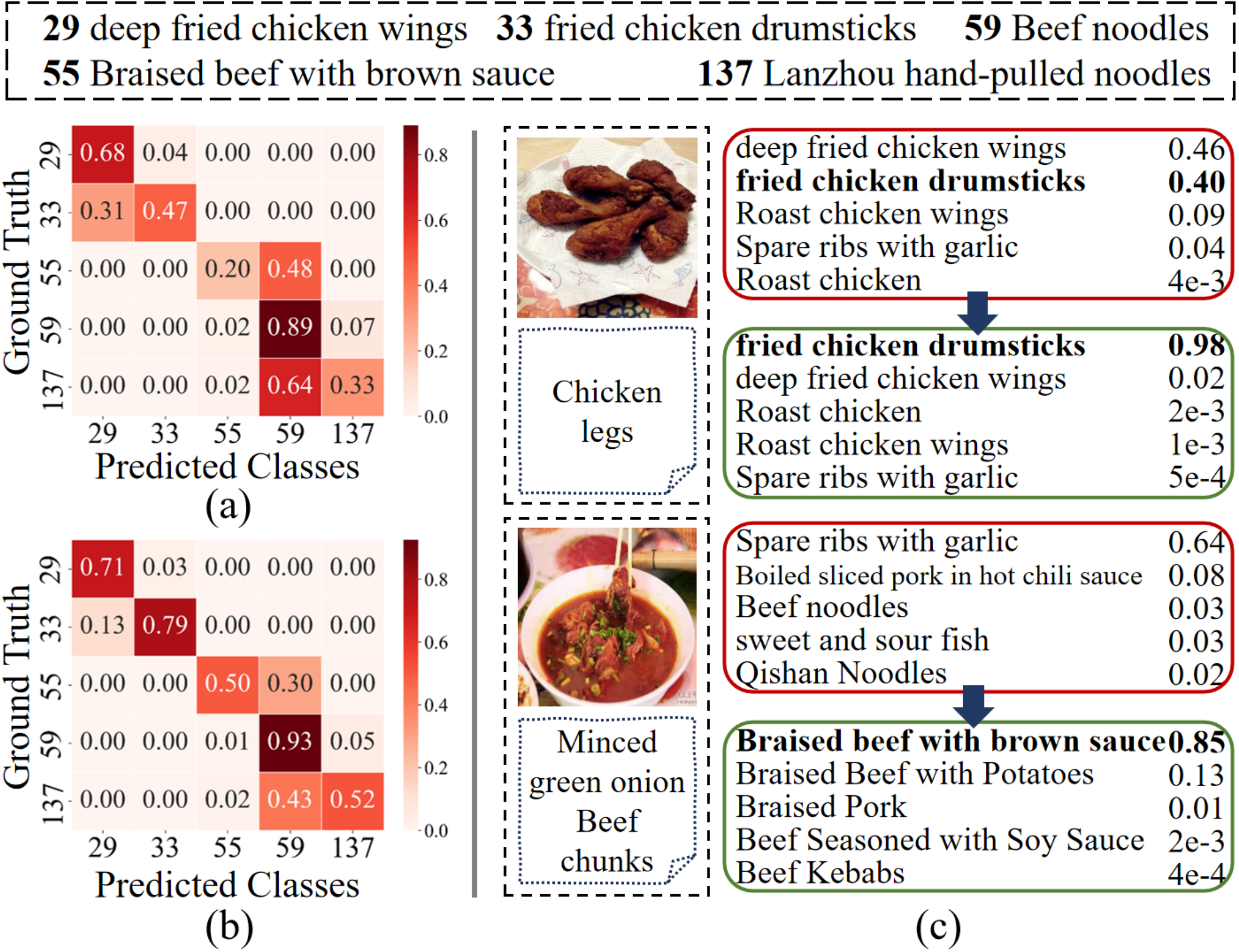}
\end{center}
% \vspace{-0.5cm}
\caption{Comparison of ViT-B/16 and SeDA on confusion matrix and randomly selected samples with Top-5 confidence scores. Red boxes represent baseline results, green boxes represent SeDA results, and ground-truth labels are highlighted in bold.}
% \vspace{-0.5cm}
\label{fig:error_analysis}
\end{figure}

\subsubsection{Effectiveness of Semantic Disambiguation} 

% 我们选择了在Baseline模型中混淆较为严重的五个类别，Baseline的混淆矩阵如图\ref{}(a)所示。ProDiffAlign对应的混淆矩阵如图\ref{}(b)所示。从结果可以明显观察到，语义相似的类别间的混淆显著减少，分类性能得到了有效提升。
% 此外，图\ref{}(c)对特定样本的预测结果进行了进一步分析。例如，对于“deep fried chicken wings”样本，在融入文本信息后，模型能够以较高的置信度预测出正确类别。这表明我们的方法能够充分利用语义相关的信息，强化对易混淆类别的区分能力。另一方面，对于包含较多背景干扰信息的样本（如“Braised beef with brown sauce”），由于原图像中存在“checkpoints”和“onion”等噪声因素，Baseline模型容易误预测为与其无关的类别（如“noodles”和“garlic”）。然而，ProDiffAlign方法能够有效提取关键特征，将类别预测准确定位到与“beef”相关的语义范围，从而显著缓解由语义歧义引发的分类错误。

We selected five categories with severe confusion in the Baseline model, as shown in Figure \ref{fig:error_analysis}(a). The corresponding confusion matrix for SeDA is shown in Figure \ref{fig:error_analysis}(b). The results clearly demonstrate that confusion between semantically similar categories is significantly reduced, leading to an effective improvement in classification performance.
Additionally, Figure \ref{fig:error_analysis}(c) presents a detailed analysis of the prediction results for specific samples. For example, in the case of the "deep fried chicken wings" sample, the model accurately predicted the correct category with high confidence after incorporating textual information. This indicates that our method effectively leverages semantic-related information to enhance the distinction between easily confused categories. On the other hand, for samples with substantial background noise (e.g., "Braised beef with brown sauce"), the Baseline model tends to misclassify them into irrelevant categories such as "noodles" or "garlic" due to noise factors like "chopsticks" and "onion" in the original image. However, SeDA successfully extracts critical features, accurately narrowing the predicted category to the semantic domain related to "beef," thereby significantly mitigating classification errors caused by semantic ambiguity.

\section{Conclusion}

% 针对多模态之间的异构性问题，我们提出了一种Diffusive Alignment between visual and textual features方法（DMANet）。DMANet通过两个阶段的扩散模型对齐，首先第一阶段学习视觉特征的结构信息，在第二阶段实现视觉特征像文本特征的逐步迁移，改善视觉特征的分布，缓解了视觉模态和文本模态之间的差异。实验结果表明，DMANet 能显著提高视觉信息的质量，优化表征的分布，提高视觉的分类性能。尽管DMANet减轻了不同模态之间的异构性问题，但是它可以在中得到改进。因此，今后的工作将着重解决这一问题。首先，探索更强的对齐策略，进一步提升视觉与文本特征对齐的精度，特别是在处理更复杂的场景和数据集时。其次，进一步探索扩散模型在跨模态对齐上的能力。

% 针对多模态之间的异构性问题，本文将扩散模型的多步去噪过程转移到视觉表示的跨模态对齐中,提出了一种渐进式扩散对齐方法（ProDiffAlign）。设计两个阶段的扩散模型对齐，首先DSCC模块学习视觉特征的结构信息，DCT模块实现视觉特征像文本特征的逐步迁移。优化视觉表征的分布，缓解了视觉模态和文本模态之间的差异，提高视觉的分类性能。尽管ProDiffAlign减轻了不同模态之间的异构性问题，但是它可以在细粒度对齐中得到改进。因此，在未来我们将继续探索扩散模型跨模态对齐的细粒度建模，进一步提升视觉与文本特征对齐的精度，特别是在处理更复杂的场景和数据集时。

% To address the heterogeneity between modalities, this paper transfers the multi-step denoising process of diffusion models to the cross-modal alignment of visual representations, proposing a semantic-space-intervened diffusive alignment method (SeDA). Using semantic space as an intermediary, a bi-stage diffusion model alignment is designed: the DSL module first captures the semantic information of visual features, and the DST module gradually translates semantic features to textual features. This method effectively projects from visual representations to textual representations.
% Despite SeDA alleviating the heterogeneity between different modalities, there remains room for improvement in fine-grained alignment. Therefore, future work will focus on exploring fine-grained modeling for cross-modal alignment using diffusion models to further enhance the alignment accuracy between visual and textual features, especially in more complex scenarios and datasets.

To address the heterogeneity between modalities, this paper transfers the multi-step denoising process of diffusion models to the cross-modal alignment of visual representations, proposing a semantic-space-intervened diffusive alignment method (SeDA). Using semantic space as an intermediary, a bi-stage diffusion model alignment is designed: the DSL module first captures the semantic information of visual features, and the DST module gradually translates semantic features to textual features. This method effectively projects from visual representations to textual representations.
Despite SeDA alleviating the heterogeneity between different modalities, there remains room for improvement in fine-grained alignment. Future work will explore fine-grained modeling with diffusion to further improve alignment accuracy, especially in complex scenarios and datasets.

% % % This approach optimizes the distribution of visual representations and mitigates differences between visual and textual modalities. 

% Despite SeDA alleviating the heterogeneity between different modalities, there remains room for improvement in fine-grained alignment. Therefore, future work will focus on exploring fine-grained modeling for cross-modal alignment using diffusion models to further enhance the alignment accuracy between visual and textual features, especially in more complex scenarios and datasets.

% To address modality heterogeneity, this paper applies the multi-step denoising process of diffusion models to cross-modal alignment, proposing a semantic-space-intervened method (SeDA). A bi-stage framework is designed: the DSL module captures visual semantics, and the DST module gradually maps them to textual features via semantic space. This enables effective projection from visual to textual representations. While SeDA alleviates modality gaps, fine-grained alignment remains a challenge. Future work will explore fine-grained modeling with diffusion to improve alignment accuracy, especially in complex scenarios and datasets.

\section*{Acknowledgments}
This work is supported in part by the Shandong Province Excellent Young Scientists Fund Program (Overseas)  (Grant no. 2022HWYQ-048).
\bibliographystyle{named}
\bibliography{ijcai25}

\end{document}